\documentclass[runningheads]{llncs}
\usepackage{graphicx}
\usepackage{booktabs}
\usepackage{amsmath}

\begin{document}
\title{Physically constrained neural networks to solve the inverse problem for neuron models.}
\titlerunning{PINN for neuron models}
%
\author{Matteo Ferrante\inst{1} \and
Andrea Duggento\inst{1}\and
Nicola Toschi\inst{2}}
\authorrunning{Ferrante et al.}
%
\institute{Department of Biomedicine and Prevention, University of Rome Tor Vergata, Rome, Italy
\and
Department of Radiology, Athinoula A. Martinos Center for Biomedical Imaging, Boston, MA, USA\\
\email{matteo.ferrante@uniroma2.it}}
\maketitle              
\begin{abstract}
Systems biology and systems neurophysiology in particular have recently emerged as powerful tools for a number of key applications in the biomedical sciences. Nevertheless, such models are often based on complex combinations of multiscale (and possibly multiphysics) strategies that require ad hoc computational strategies and pose extremely high computational demands. Recent developments in the field of deep neural networks have demonstrated the possibility of formulating nonlinear, universal approximators to estimate solutions to highly nonlinear and complex problems with significant speed and accuracy advantages in comparison with traditional models. After synthetic data validation, we use so-called physically constrained neural networks (PINN) to simultaneously solve the biologically plausible Hodgkin-Huxley model and infer its parameters and hidden time-courses from real data under both variable and constant current stimulation, demonstrating extremely low variability across spikes and faithful signal reconstruction. The parameter ranges we obtain are also compatible with prior knowledge. We demonstrate that detailed biological knowledge can be provided to a neural network, making it able to fit complex dynamics over both simulated and real data.

\end{abstract}

\section{Introduction}

Physiological systems modeling in general and the novel paradigms of systems biology and systems physiology, in particular, have recently emerged as powerful tools for a number of key applications in the biomedical sciences, such as patient stratification, the discovery of disease mechanisms, and \textit{in silico} drug design. Nevertheless, systems biology and systems pharmacology models are often based on complex combinations of multiscale (and possibly multiphysics) models that require \textit{ad hoc} computational strategies and pose extremely high computational demands. In this context, recent developments in the field of deep neural networks have demonstrated the possibility of formulating nonlinear, universal approximators to estimate solutions to highly nonlinear and complex problems with significant speed and accuracy advantages in comparison with traditional models. Recently, a new approach to solve partial differential equations has emerged in the form of so-called Physically Constrained- (or Physics- Informed-) Neural Networks (PINN)  \cite{raissi2017physics,raissi2017physics2}. In PINNs, a feed-forward network is used to find the approximation of  a function that satisfies a set of differential equations and boundaries as well as initial conditions. This “soft constrained” approach is based on satisfying analytical constraints by minimizing the L2 norm coupled with the boundary and initial  conditions in the loss function. Initial applications of PINNs have been seen to achieve results that are indistinguishable (to machine precision) to conventional numerical integrators, with the advantage of an extremely flexible and fast framework with arbitrary large representational capability  \cite{2021NatRP...3..422K}. Moreover, PINNs are mesh-free and can also be employed to solve inverse problems by providing sparse measurements to the network and jointly optimizing network parameters and model parameters to fit real-world data, hence effectively solving the model system while simultaneously parameterizing it to best describe a given observation.
Within neuroscience applications of systems biology, ac- curate modeling of neuronal networks is key in order to better understand the intricacies of brain functions. In this context, previous work has relied either on massively parallel computer simulations accessible to only a few researchers in the world  \cite{MARKRAM201139}, or on simplified models to be able to simulate larger populations \cite{simplespikingmodel}.  While the integrate and fire framework \cite{Burkitt2006-hs} has been widely employed to approximate neuronal dynamics, it lacks the physiological detail to truly understand the impact of neuronal dynamics in applications such as neurostimulation \cite{kamimura}, learning, and plasticity \cite{Toschi2009-te}. Much more detailed biological models like the Hodgkin-Huxley (HH) model \cite{HODGKIN1952} are able to take into account the ionic current of Na, K, and other species through the membrane of the neuron; however, a large-scale simulation of HH neurons presents considerable computational complexities. Most importantly, however, employing the HH model for neuron-level neuroscience requires knowledge about model parameters that cannot be directly measured \cite{Doruk2019-wh}. 

Prior work \cite{denseparameter} h] has shown that NNs are able to recover parameters of simulated neuronal spiking using the Fitzhugh- Nagumo (FN) model (i.e., a linearized version of the HH model). The aim of this paper was 1) to design a PINN architecture able to infer hidden parameters in the physiologically accurate HH neuron model, 2) to validate our framework on simulated, ground-truth data, and 3) to demonstrate a proof-of-concept application employing publicly available neurophysiological measurements from single neurons obtained from the Physionet repository.

\section{Theory}
In this paper, we explore the performance of PINNs in estimating neuron model parameters on both synthetic and real data in both the FN \cite{FITZHUGH1961445,FitzHugh1955} and the more realistic but nonlinear HH model.  

\begin{figure}[t]
\centering\includegraphics[width=0.9\textwidth,]{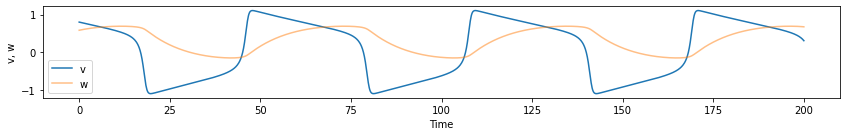}
	\caption{
		\label{fig:} Example of an FN model realization. The parameters of the simulation are $a=-0.3$, $b=1.2$, $I=0.28$, $\tau=20$.}
\end{figure}

\subsection{Hodgkin-Huxley model}
The HH model is the most established neuron modeling framework. It explicitly accounts for the biological flux of the main chemical species through the cell membrane. The lipid bilayer is modeled as a capacitance  $C_m$, while the voltage-gated ion channels are modeled as conductance $g_{\text{ionic}}$. When mimicking a neurophysiology experiment, one is interested in inferring the membrane potential $V_m$ as a function of time in response to an external current stimulus $I$.
The current flowing through the membrane can be written as $ I_{c}=C_{m}{\frac {{\mathrm {d} }V_{m}}{{\mathrm {d} }t}}$ and the ion channel currents as $I_{i}={g_{i}}(V_{m}-V_{i})$, where $V_i$ is the reversal potential for the $i$th ionic species, namely the potential at which there is no net flow of that particular ion through the channel.
The HH model includes ion-specific conductance and models membrane potential through a system of four ordinary differential equations

\begin{equation}
\begin{split}
     I=&C_{m}\dot{V_{m}}+
     {\bar {g}}_{\text{K}}n^{4}(V_{m}-V_{K})\\
     &+{\bar{g}}_{\text{Na}}m^{3}h(V_{m}-V_{Na})+{\bar {g}}_{l}(V_{m}-V_{l})\\
     \dot{n} &=\alpha _{n}(V_{m})(1-n)-\beta _{n}(V_{m})n \\
     \dot{m} &=\alpha _{m}(V_{m})(1-m)-\beta _{m}(V_{m})m \\
     \dot{h} &=\alpha _{h}(V_{m})(1-h)-\beta _{h}(V_{m})h 
\end{split}
\end{equation}
where $\alpha$ and $\beta$ are empirical functions of the membrane and reverse potentials. In this paper, this system can be evolved numerically for different input current waveforms to generate synthetic data from which we attempt to recover hidden model parameters through the use of PINNs. System evolution can also be performed through freely available software packages such as NEST \cite{Gewaltig:NEST} or BRIAN \cite{Stimberg2019}.

\subsection{Fitzhug-Nagumo model}
The FN model simplifies the HH model to two differential equations by only taking into account the faster components (i.e. the membrane potential) and a slow recovery function responsible for the refractory period of the neuron just after the spiking phenomenon as follows:

\begin{equation}
    \begin{split}
    \dot{v}=v-v^{3}-w+RI_{ext} \\
    \tau {\dot {w}}=v+a-bw
    \end{split}
\end{equation}

where $v$ is the membrane potential, $w$ is a recovery variable with a characteristic time $\tau$.  $a$ and $b$ are extra parameters that modulate the evolution of $w$ and $I$ is the current that passes through the membrane with resistance $R$.

\subsection{PINNs}
Among the many applications of modern neural networks, their use to solve differential equations under analytical constraints or even solve possibly ill-formulated inverse problems is still relatively unexplored in the biological domain. In the latter case, the goal is to estimate an optimal set of unknown parameters $\lambda$ given the governing equations and real world measurements of the phenomena modeled through those equations.
The goal is to estimate a function $u(t)$ that satisfies some physical constraints in the form of differential equations parameterized by $\mathbf{\lambda}$

\begin{equation}
    f(\mathbf{t},u,\dot{u},\mathbf{\lambda})=0
\end{equation}
 where $t \in \Omega$,
it is possible to train a neural network to approximate the solution by computing the $L_2$ norm of the difference between this constraint and a subset of data points by taking advantage of automatic differentiation to compute first and higher order derivatives of variables, in the same way they are computed as in a standard neural network.
In this way, the network parameterized by $\theta$ is trained to approximate $\tilde{u}(t,\theta,\lambda) $ by optimizing a loss function that is usually composed of three terms:

\begin{equation}
    \mathcal{L}_f=\frac{1}{\tau_f}\sum_{t \in \tau_f} \lVert f(\mathbf{t},\frac{\partial u}{\partial t},..\mathbf{\lambda})  \rVert^2_2
\end{equation}

\begin{equation}
    \mathcal{L}_b=\frac{1}{\tau_b}\sum_{t \in \tau_b} \lVert \mathcal{B}(\tilde{u}(t,\lambda)\rVert^2_2 
\end{equation}

\begin{equation}
    \mathcal{L}_m=\frac{1}{\tau_m}\sum_{t \in \tau_f} \lVert (\tilde{u}(t_i,\lambda) - u(t_i)  \rVert^2_2
\end{equation}

\begin{figure}[t]
\centering\includegraphics[width=0.9\textwidth]{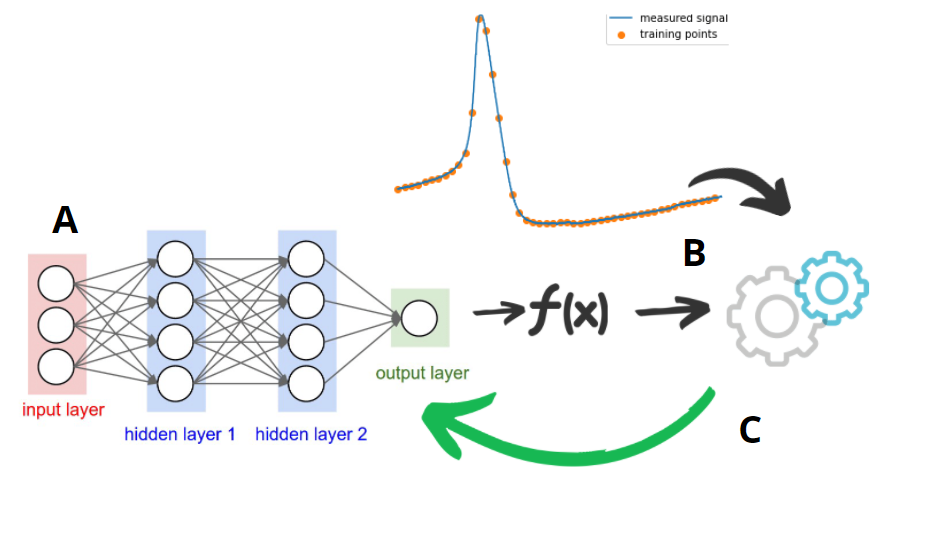}
	\caption{
		\label{fig:pinn} Example of how a physically constrained neural network solves an inverse problem. 
		A. A set of in-domain and boundary points are passed to the network that outputs an estimate of the expected function values at those points.
		B. The loss is computed taking into account how well the differential equation is satisfied along with its initial and boundary condition. In the case of inverse problems, a term that measures the discrepancy from measured data is added to the loss function to infer the model parameters
		C. Gradient descent optimizes the network weights $\theta$ and the model parameters $\lambda$ to minimize the total loss function and fit the data.}
\end{figure}

The $\mathcal{L}_f$ term anchors the solution to the data points provided in the training set  $\tau_f$.
The second term $\mathcal{L}_b$ accounts for the boundary and initial condition, effectively penalizing trivial solution.
The third term is only included when solving an inverse problem, i.e. real-world data are available and the goal is to estimate hidden parameters in the governing equations. In this case, the term $\mathcal{L}_m$ is included in the total loss function in order to force the learning process to jointly  $\theta$ and $\lambda$ and therefore estimate both the solution to the governing equations while simultaneously parameterizing the latter  equations to optimally describe observations. This approach is especially 
powerful in both forward and inverse problems including all types of differential equations.
Specifically, the network is provided with a set of measured points and boundary conditions as well as with a set of points where the output of the network must satisfy the constraints imposed by the differential equation and boundary conditions at the same time. The parameters are updated using gradient descent until these criteria are satisfied within a tolerance similar to machine like precision as described in Fig \ref{fig:pinn}.

\section{Methods}

\subsection{Validation}
To test the ability of the PINN approach to infer models for neuronal dynamics, we integrated the FN model using the scipy \cite{2020SciPy-NMeth} odeint function (300 timepoints with initial parameters set as $a=-0.3$, $b=1.2$, $I=0.28$, $\tau=20$), and applied our PINN to solve the inverse problem to explore the goodness of fit as well as reproducibility. Following synthetic validation, we applied our approach to solve the inverse problem using both the FN and the HH model based on publicly available data measured from the squid giant axon (\url{https://archive.physionet.org/physiobank/database/sgamp/}), both in the case of zero input current and in the case of variable input currents. We perform multiple fits of different single spikes with zero input current to explore the variability of the inferred parameters both in the FN and HH models. 

\subsection{PINNs}
Experiments were performed using deepxde \cite{RAISSI2019686} with Pytorch backend.
In our first experiment, we defined a feed-forward network with 3 hidden layers with 40 neurons each and a final layer with 2 outputs (activation function: "tanh") to estimate the $v$ and $w$ components of the FN model from simulated data.  The network was trained for 60000 epochs with the Adam optimizer with a learning rate of 0.001 and constrained by 30 uniformly sampled values from the simulated membrane along with suitable initial conditions. When using real data, a number of single spikes were fitted using both models (FN: 3 layers/50 neurons/2 output neurons, 25000 epochs; HH: 3 layers/60 neurons/2 output neurons, 23000 epochs) to investigate stability and variability.

\section{Results}


\begin{figure}[h!]
\centering\includegraphics[width=0.95\textwidth]{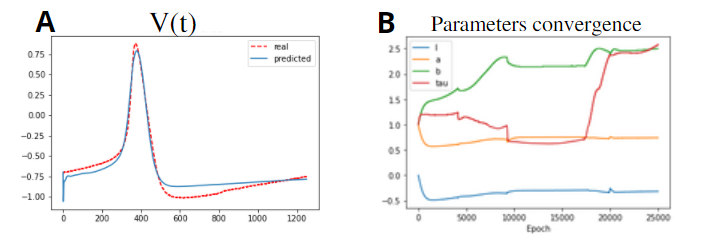}

	\caption{
		\label{fig:} Example  results (single spike) using the FN model to fit real data.
		A. Approximation of membrane potential from network
		B. Evolution of the model parameters across iterations}
\end{figure}

\begin{figure}[h!]
\centering\includegraphics[width=0.9\textwidth]{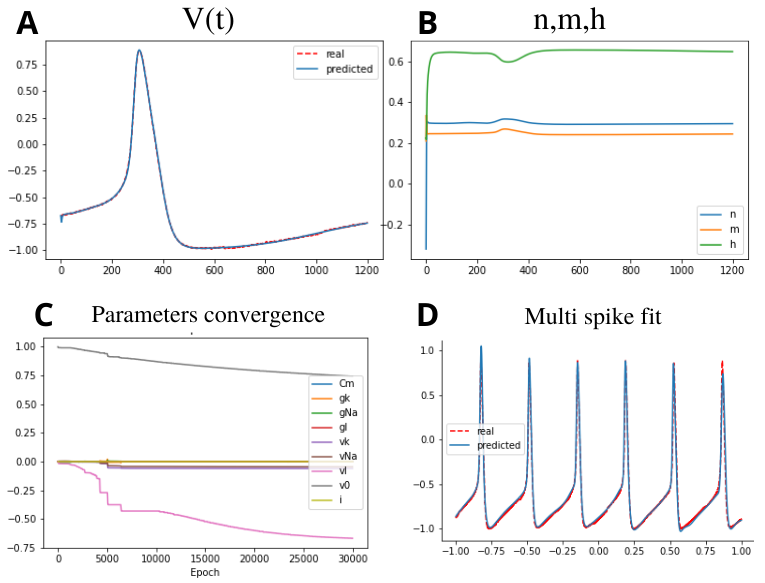}
	\caption{
		\label{fig:} Example results (single spike and spike train) using the HH to fit real data.
		A. Approximation of membrane potential from network
		B. Approximation of hidden n,m,h components (ionic currents)
		C. Evolution of the model parameters across iteration
		D. Fit of multiple subsequent spikes using the HH (rescaled to [-1,1])
		}
\end{figure}

\begin{figure}[h!]
\centering\includegraphics[width=0.9\textwidth]{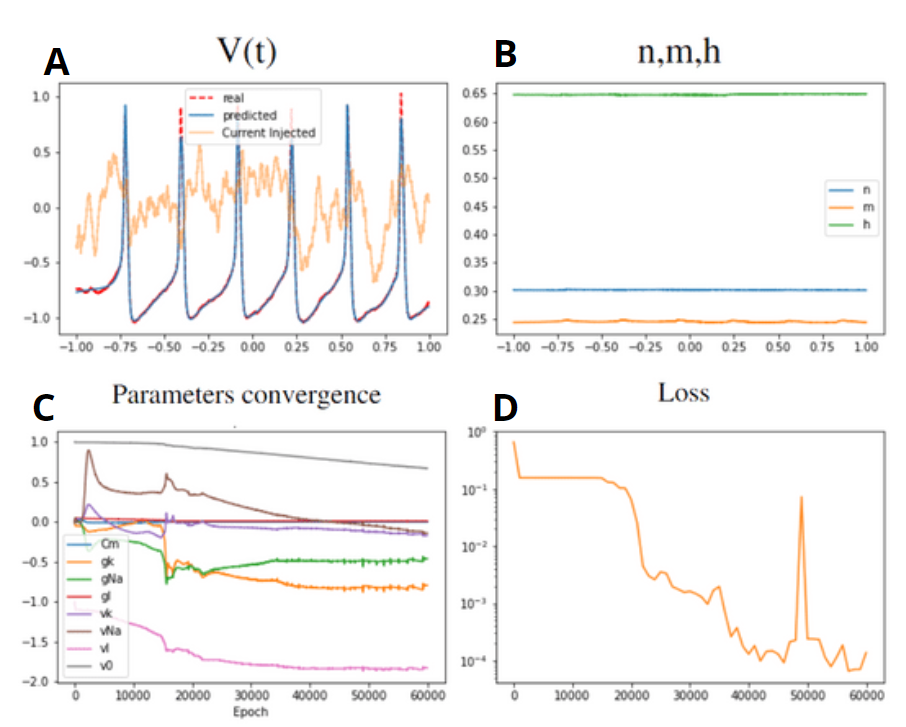}
	\caption{
		\label{fig:time} Example results on a train of spikes under time-varying input current using the HH model.
		A. Approximation of membrane potential from network
		B. Approximation of hidden n,m,h components (ionic currents across the cell channels)
		C. Evolution of the model parameters across iterations
        D. Fit of multiple subsequent spikes using this model (rescaled to [-1.,1.])
		}
\end{figure}

\begin{figure}[h]
\centering\includegraphics[width=0.95\textwidth]{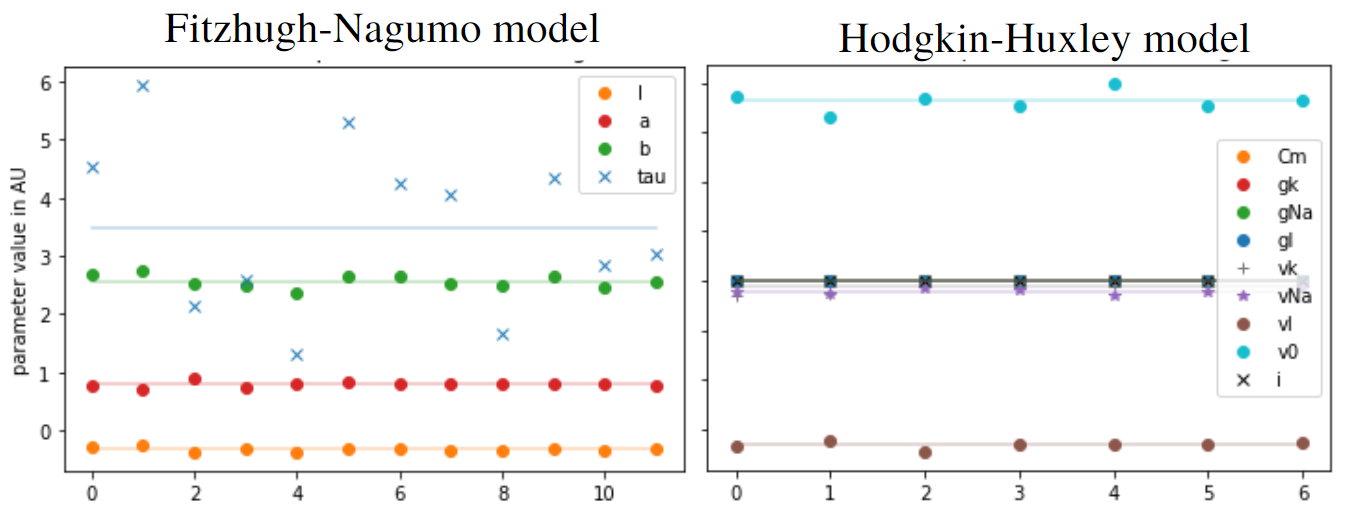}
\caption{\label{fig:parameters} Inferred parameters across real data spikes for the FN (left) and HH (right).}
\end{figure}

Parameters inferred from the synthetic data were $I= 0.279$ $a= -0.319$ $b= 1.26$ $\tau= 19.7$., i.e. extremely close to the original parameters used to generate synthetic data. Using real data, we first fitted 12 single spikes with the FN model (Table \ref{table:fitz}) and 7 single spikes with the HH model (Table \ref{table:hodgkin}) to explore the variability of the inferred parameters across subsequent spikes. In the former, inferred $I$, $a$ and $b$ are extremely stable across spikes,  $\tau$ shows higher variability. In the latter, all parameters are inferred have an extremely narrow variability range across spikes (Fig. \ref{fig:parameters}). 
Fig. \ref{fig:time} show the results of fitting multiple spikes into a time-varying current frame. We qualitatively find the same order of magnitude when comparing the conductance and the resting potential between them to the ones that we previously found with the analysis with zero input current.

\begin{table}[h!]
\large
\centering
\begin{tabular}{@{\hskip 12pt}l@{\hskip 12pt}l@{\hskip 12pt}l@{\hskip 12pt}l@{\hskip 12pt}l@{\hskip 12pt}l@{}}
\toprule
   & mean    & $\sigma$  & $\sigma/\mu$   & min    & max    \\ \midrule
\textbf{I}   & -0.3264 & 0.0011 & 3.3e-3 & -0.377 & -0.258 \\
\textbf{a}   & 0.7921  & 0.0020 & 2.5e-3 &  0.72   & 0.9    \\
\textbf{b}   & 2.565   & 0.0119 & 4.6e-3 & 2.38   & 2.75   \\
$\boldsymbol{\tau}$ & 3.493   & 2.132  & 0.61  &  1.3    & 5.93   \\ \bottomrule
\end{tabular}

\caption{\label{table:fitz}FN model parameters inferred by the PINN}
\end{table}

\begin{table}[h!]
\centering
\large
\begin{tabular}{@{}l@{\hskip 12pt}l@{\hskip 12pt}l@{\hskip 12pt}l@{\hskip 12pt}l@{\hskip 12pt}l@{}}
\hline
    & mean     & $\sigma$  & $\sigma/\mu$     & min       & max \\ \midrule
\textbf{Cm}  & 8.85e-6  & 3.7e-10 &  4.1e-5   & -7.32e-6  & 4.76e-5  \\
\textbf{gk}  & 6.39e-5  & 2.73e-8 &  3.9e-4   & -0.000121 & 0.000395 \\
\textbf{gNa} & -5.96e-5 & 1.19e-8 &  3.2e-4   & -0.000338 & 9.61e-05 \\
\textbf{gl}  & 2.51e-6  & 1e-10   &  3.9e-5   & -8.68e-6  & 2.1e-5   \\ 
\textbf{vk}  & -0.0241  & 0.0005  &  2.1e-2   & -0.06     & -0.0071  \\
\textbf{vNa} & -0,041   & 0.0001  &  2.4e-3   & -0.0553   & -0.0276  \\
\textbf{vl}  & -0.662   & 0.0002  &  3.0e-4   & -0.689    & -0.647   \\
\textbf{v0}  & 0.726    & 0.0016  &  2.2e-3   &  0.662     & 0.795    \\
\textbf{i}   & 0.       & 0.      & 0.        & 0.       & 0.\\ \bottomrule
\end{tabular}

\caption{\label{table:hodgkin} HH model parameters inferred by the PINN}
\end{table}

\section{Discussion and Conclusion}
In this work, we showed, through a proof-of-concept experiment on neuronal spiking data, how physically constrained neural networks can quickly and accurately solve problems in  domains such as biology and neuroscience, where a set of differential equations describes complex dynamics. Interestingly, when solving the HH model and simultaneously inferring its parameters even from sparse real data, our results are in accordance with values reported elsewhere (when appropriately rescaled to revert normalization operations performed in our PINN) \cite{estimation}. In detail, $g_k$, $g_Na$, $v_k$ and $v_{Na}$ have comparable orders of magnitude with previous estimates,while $g_l$ is roughly ten times smaller. $v_l$ was estimated to be much greater, which can probably be ascribed to the fact the membrane was pierced with an electrode for measurement (i.e. increasing leakage). In summary, PINNs can easily solve inverse problems by formulating them as forward ones, enabling the inference of hidden variables that are not usually observable and thus arduous to extract with more conventional approaches. We demonstrate that detailed biological knowledge can be provided to a neural network, making it fit complex dynamics over both simulated and real data.

\section*{Acknowledgements}

{\footnotesize
Part of this work is supported by the EXPERIENCE project (European Union’s Horizon 2020 research and innovation program under grant agreement No. 101017727)

The Titan V GPUs employed in this research were generously donated to NT by NVIDIA.

Matteo Ferrante is a PhD student enrolled in the National PhD in Artificial Intelligence, XXXVII cycle, course on Health and life sciences, organized by Università Campus Bio-Medico di Roma.

}

%
%
%
%
%
%
\bibliographystyle{splncs04}
\bibliography{paper}
\end{document}